%% file: acl_latex.tex
\documentclass[11pt]{article}
\usepackage[english]{babel}
\usepackage[final]{acl}

\usepackage{times}
\usepackage{latexsym}

\usepackage[T1]{fontenc}

\usepackage[utf8]{inputenc}

\usepackage{microtype}
\usepackage{enumitem}
\usepackage{inconsolata}
\usepackage{subcaption}
\usepackage{graphicx}
\usepackage{booktabs}
\usepackage{multirow}
\usepackage{tabularx}

\title{DRAGOn: Designing RAG On Periodically Updated Corpus}
\author{
 \textbf{Fedor Chernogorskii\textsuperscript{2,1}},
 \textbf{Sergei Averkiev\textsuperscript{1}},
 \textbf{Liliya Kudraleeva\textsuperscript{3}},\vspace{-10pt}
 \\ \\ 
 \textbf{Zaven Martirosian\textsuperscript{1,4,8}},
 \textbf{Maria Tikhonova\textsuperscript{1,5}},
 \\
 \textbf{Valentin Malykh\textsuperscript{6,3,7}},
 \textbf{Alena Fenogenova\textsuperscript{1,5}}
\\
 \textsuperscript{1}SberAI,
 \textsuperscript{2}MBZUAI,
 \textsuperscript{3}ITMO,
 \textsuperscript{4}MISIS,
 \textsuperscript{5}HSE University,
 \textsuperscript{6}MWS AI,
 \textsuperscript{7}IITU,
 \textsuperscript{8}YSDA
\\
 \small{
   \textbf{Correspondence:} \href{mailto:fechernogor@gmail.com}{fechernogor@gmail.com}
 }
}

\begin{document}

\selectlanguage{english}
\maketitle             

\begin{abstract}
This paper introduces \textbf{DRAGOn}, method to design a RAG benchmark on a regularly updated corpus. It features recent reference datasets, a question generation framework, an automatic evaluation pipeline, and a public leaderboard.
Specified reference datasets allow for uniform comparison of RAG systems, while newly generated dataset versions mitigate data leakage and ensure that all models are evaluated on unseen, comparable data.
The pipeline for automatic question generation extracts the Knowledge Graph from the text corpus and produces multiple question-answer pairs utilizing modern LLM capabilities.
A set of diverse LLM-as-Judge metrics is provided for a comprehensive model evaluation.
We used Russian news outlets to form the datasets and demonstrate our methodology. We launch a public leaderboard to track the development of RAG systems and encourage community participation.
\end{abstract}

\section{Introduction}
\label{sec:intro}

Retrieval-Augmented Generation (RAG) has become a powerful tool for enhancing the domain adaptation and factuality of large language models (LLMs) by incorporating external knowledge retrieved at inference time. This approach enables more up-to-date and grounded responses without the need for costly re-training. As RAG-based systems expand to applications such as open-domain QA, customer support, and enterprise search, their standardized evaluation remains a challenge. It may be unclear whether strong performance of a system is due to the quality of its retriever-generator pipeline or because the underlying LLM has been exposed to portions of the test data during training. It is possible that a static benchmark will become contaminated over time.

\begin{figure}[ht!]
\centering
\includegraphics[width=0.4\textwidth]{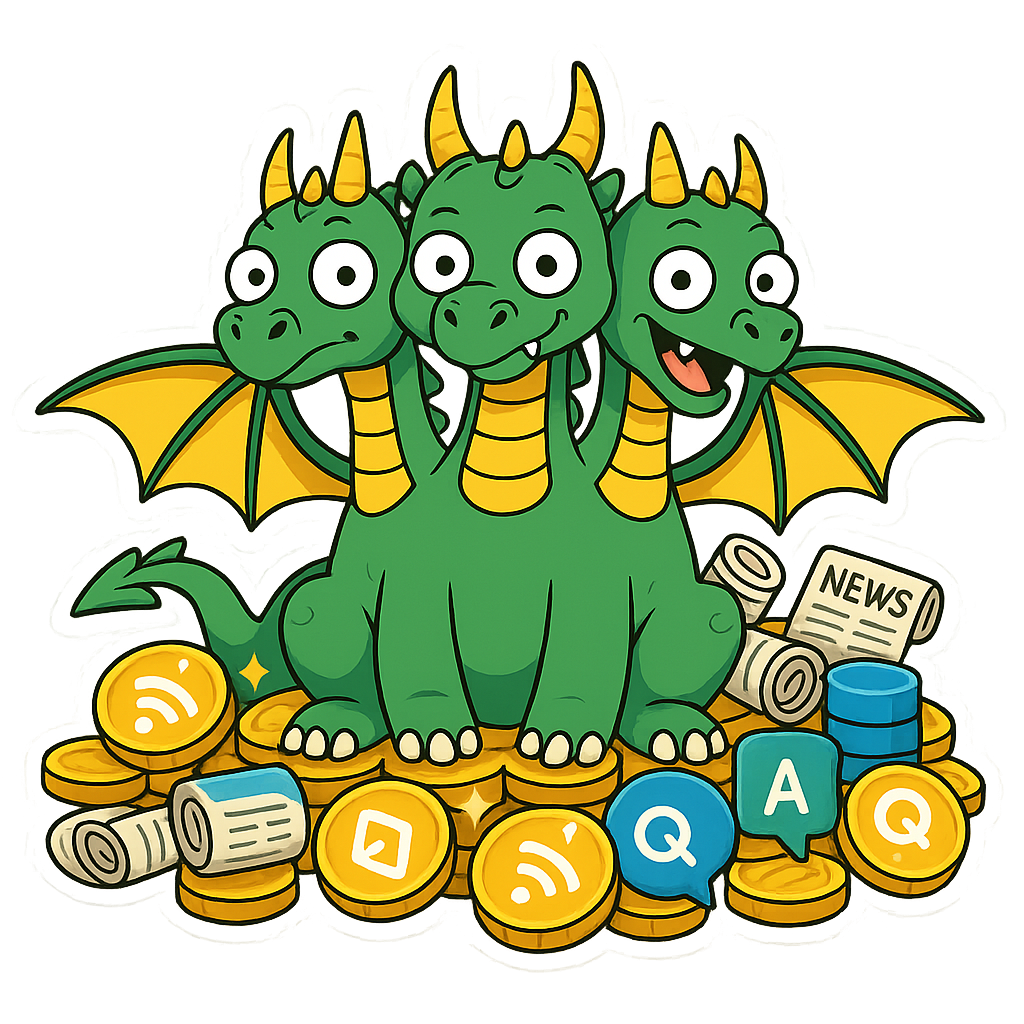} 
\caption{The DRAGOn logo.}\label{fig:dragon}
\end{figure}

Several existing RAG evaluation frameworks~\cite{es2024ragas,lyu2025crud} provide pipelines for automatic generation of question-answer pairs, typically assuming that users will build their own evaluation datasets. While this enables domain-specific benchmarking, it is labor-intensive and difficult to maintain. Furthermore, when the retrieval corpus is continuously updated during deployment, results may become non-reproducible, as the model may face a different knowledge distribution than the one used during its initial evaluation.

In this work, we introduce \textbf{DRAGOn: Designing RAG On Periodically Updated Corpus} 
a novel methodology that reflects realistic usage patterns by leveraging a regularly updated knowledge base. We also release a benchmark built on current news sources using this methodology, which can be readily adapted to other document domains, such as scientific papers or court decisions. To foster transparency and community engagement, we publicly release an  evaluation framework which comprises the codebase for automatic question generation, evaluation scripts, and a dynamic leaderboard to track progress on RAG-based systems in Russian. Although the benchmark targets Russian, the framework is potentially extendable to other languages and multilingual scenarios, making it broadly applicable. \textbf{Our contributions are as follows:}


    (i) We propose \textbf{DRAGOn}\footnote{The \href{https://www.youtube.com/watch?v=CExzet4ABG8}{video demonstration} of the evaluation tool is available on YouTube.}
    the methodology to develop a RAG benchmark with a regularly updated knowledge base, designed to evaluate RAG systems in a dynamic setup; we develop a benchmark on Russian news corpora as a reference for the proposed methodology.
    
    (ii) We release an open-source evaluation framework
    \footnote{The framework is released under the MIT license: \url{https://github.com/RussianNLP/DRAGON}}
    comprising a reusable question generation pipeline and evaluation scripts, enabling reproducible experimentation and easy integration of new models and retrieval components. By design, it can potentially be adapted to other languages and multilingual settings, broadening its applicability beyond Russian.
    
    (iii) We launch a regularly updated public leaderboard\footnote{\url{https://huggingface.co/spaces/ai-forever/rag-leaderboard}}
    for recurrent evaluation
    to support reproducible and community-driven research.

\section{Related Work}
\label{sec:related_work}

Evaluating retrieval-augmented generation (RAG) systems poses unique challenges, as it requires datasets that jointly assess both the retrieval and generation components. Constructing such benchmarks is costly and time-consuming because it involves curating large collections of text–question–answer triplets. To alleviate this, several works have explored synthetic data generation to automate question and answer creation~\cite{es2024ragas,lyu2025crud}, often leveraging domain-specific pipelines or knowledge graphs for better control over content and difficulty.

Early RAG benchmarks such as KILT~\cite{petroni2021kilt} unified multiple English-language datasets over a fixed Wikipedia snapshot, emphasizing source attribution and retrieval grounding. More recent efforts have extended the evaluation to multi-turn conversational and reasoning-intensive scenarios, as seen in mtRAG~\cite{katsis2025mtrag} and RAD-Bench~\cite{kuo2025radbench}. The CRAG benchmark~\cite{yang2024crag} further focuses on factual consistency, capturing five key aspects of RAG system behavior. Complementarily, RAGAS~\cite{es2024ragas} provides a reference-free evaluation framework measuring context relevance, faithfulness, and answer completeness, and offers an open-source API for reproducible benchmarking.

Dynamic and time-sensitive evaluation has emerged as another important dimension. RealTime QA~\cite{kasai2023realtimeqa} introduced a benchmark for evaluating systems on continuously evolving information sources, reflecting real-world deployment settings. The news domain, with its frequent updates and temporal drift, has thus become a popular testbed for such studies~\cite{tang2024multihoprag,chen2024benchmarking}.

Despite these advances, the field still lacks a universal, contamination-free, and continuously updated benchmarks~\cite{white2024livebench}. This gap hinders fair and reproducible comparison across RAG systems and motivates the development of dynamic, standardized evaluation resources.

To address this need, we present \textbf{DRAGOn} -- a methodology to develop dynamic, regularly updated benchmarks for RAG systems based on real-world, shifting corpora.

\section{Benchmark Design}
\label{sec:system_demo}

\begin{figure*}[tbh!]
    \centering
    \includegraphics[width=0.9\textwidth]{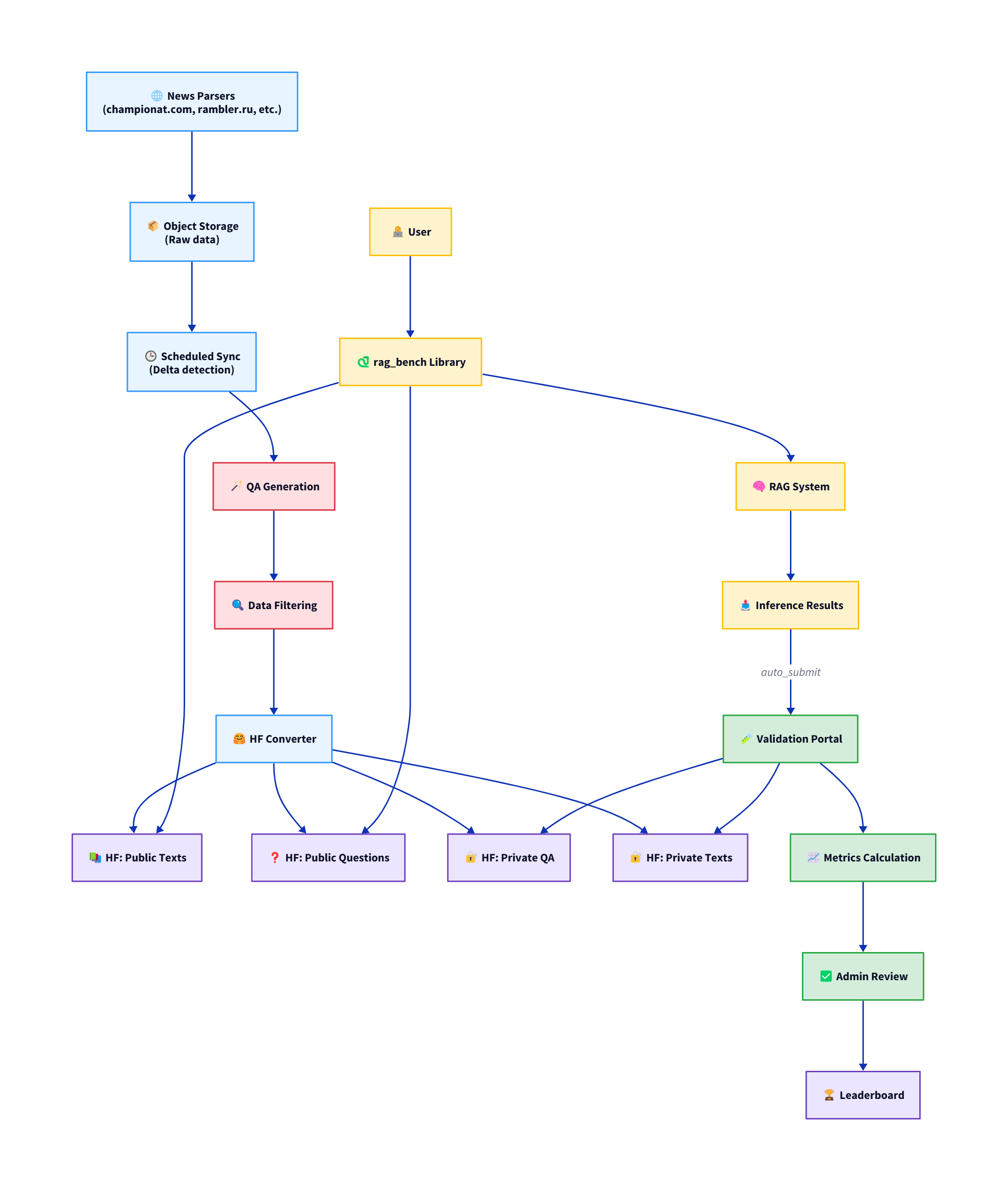}
    \caption{Architecture of the benchmark system based on DRAGOn. All datasets are versioned and uploaded to Hugging Face with incrementally updated revision numbers. This versioning mechanism ensures reproducibility and provides users with stable snapshots for further experimentation.}
    \label{fig:dragon-pipeline}
\end{figure*}

Using our methodology, one can develop a benchmark designed to evaluate RAG systems in a dynamically evolving news domain. The benchmark's architecture prioritizes modularity, automation, and reproducibility while addressing the core challenges~\cite{yu2024evaluation} in the RAG evaluation landscape, such as the temporal aspects of information, the vast and dynamic sources of knowledge, and the factuality and faithfulness in generation.
The entire pipeline of the benchmark architecture is shown in Fig.~\ref{fig:dragon-pipeline}. Below, each step is described in more detail.

\textit{Data Acquisition and Processing:}
\label{subsec:raw_data}
We maintain a dedicated set of parsers which periodically crawl a selection of news sources recognized as popular news websites in Russia on a daily basis. The parsed content is synchronized with our storage. To avoid redundancy and ensure incremental updates, a scheduled automated job identifies differences with the previous dataset revision and extracts updated segments for downstream processing. 

This design ensures that the benchmark reflects evolving real-world distributions and mitigates the risks of overfitting to static datasets. The pipeline further ensures that newly surfaced topics and entities from the news stream are constantly incorporated into the benchmark.

\textit{QA Dataset Formation:}
\label{subsec:hf_dataset}
The process of creating questions and answers based on the updated increment of the news data is described in detail in Sec.~\ref{sec:methodology}. The pipeline transforms the generated QA pairs into several HF datasets, which form the core of the benchmark:

\begin{itemize}
    \item \textit{Public Texts}: Contains cleaned source documents. Each item is assigned a \texttt{public\_id} to enable matching without exposing the true internal IDs.
    \item \textit{Public Questions}: Contains only questions, indexed via \texttt{public\_id} to obfuscate alignment and encourage retrieval.
    \item \textit{Private Texts Mapping}: Used only for evaluation purposes. It contains internal \texttt{id}s and the corresponding \texttt{public\_id}s to enable accurate mapping during metric computation.
    \item \textit{Private QA}: Provides canonical ground-truth answers for generative evaluation.
\end{itemize}

In addition to these main datasets, we provide a separate set of \textit{Sandbox Datasets} with the exact same structure as the main ones. All four sandbox datasets are fully public. Their purpose is twofold: (1) to transparently demonstrate the full structure and intended usage of the benchmark, and (2) to allow users to validate their RAG systems locally without submitting results to the validation portal.

These sandbox datasets can be evaluated using the \texttt{rag\_bench} client library, which supports the same retrieval and generative metrics as those used by the official validation portal (except for judgment-based metrics). This enables convenient local experimentation, debugging, and reproducibility.

\paragraph{User Experience}
\label{subsec:user}
To facilitate seamless evaluation for users, we provide a PyPi-hosted Python library \texttt{rag\_bench}, which offers an interface to:
\textit{Fetch} the latest version of the public datasets by dynamically resolving the latest Hugging Face revision; \textit{Observe} the RAG system baseline, which can be adopted for the target one; \textit{Evaluate} RAG system and package results for submission;  \textit{Submit} results via API to our evaluation portal; \textit{Calculate} retrieval and generative metrics locally using the sandbox datasets.

User workflow includes loading public data, applying a custom RAG pipeline, and collecting results in the following form:
{\small
\begin{verbatim}
{   
    "0": {
        "found_ids": [17, 69, 69, 22, ...],
        "model_answer": "Answer"
    }, 
    ...,
}
\end{verbatim}
}
These results encode both the retrieved \texttt{public\_id}s and the generated answers, decoupling the user’s model output from any private evaluation artifacts. This separation allows for secure evaluation without exposing ground-truth data.

\paragraph{Validation Portal}
\label{subsec:val_portal}

Submitted results then are sent to the \textit{Validation Portal} --- a Flask-based backend with a Single Page Application written in Vue as a frontend that performs secure evaluation using the private datasets. The portal evaluates submissions using private datasets and prepares evaluation results for admin approval before publishing.
Importantly, users submit only their results --- all ground-truth data remains internal.

\paragraph{Leaderboard and Auto-Evaluation}
\label{subsec:leaderboard}

A Hugging Face Gradio Space serves as the public {Leaderboard}. The results are committed in a version-controlled \texttt{results.json} file, automatically updated by the validation portal upon approval.

To reduce latency and improve benchmarking coverage, we support automatic evaluation for selected pre-approved baselines, which include several popular LLMs and retrieval embedding models. The results are computed via the same \texttt{rag\_bench} client.

\subsection{Versioning Strategy}
\label{subsec:versioning}

Given the dynamic nature of the benchmark, versioning plays a critical role in ensuring meaningful comparisons. Each evaluation result is tied to a specific dataset revision. On the leaderboard, users can view results for a single dataset version or toggle an ``Actual Versions'' mode to aggregate results across recent revisions.

Dataset versioning is performed automatically based on the last available version on Hugging Face. The version number follows a semantic format, e.g., \texttt{1.10.0}. For each new release, the middle segment of the version is incremented, resulting in a new version such as \texttt{1.11.0}, which is then uploaded to Hugging Face. This approach ensures consistent, chronological dataset updates while preserving backward compatibility for previously published results.

Note that sandbox datasets are not updated on a regular basis. They serve as a static reference set for demonstration and local validation purposes.

\section{Dataset Generation}
\label{sec:methodology}
\begin{figure}[tbh!]
    \centering
    \includegraphics[width=\columnwidth]{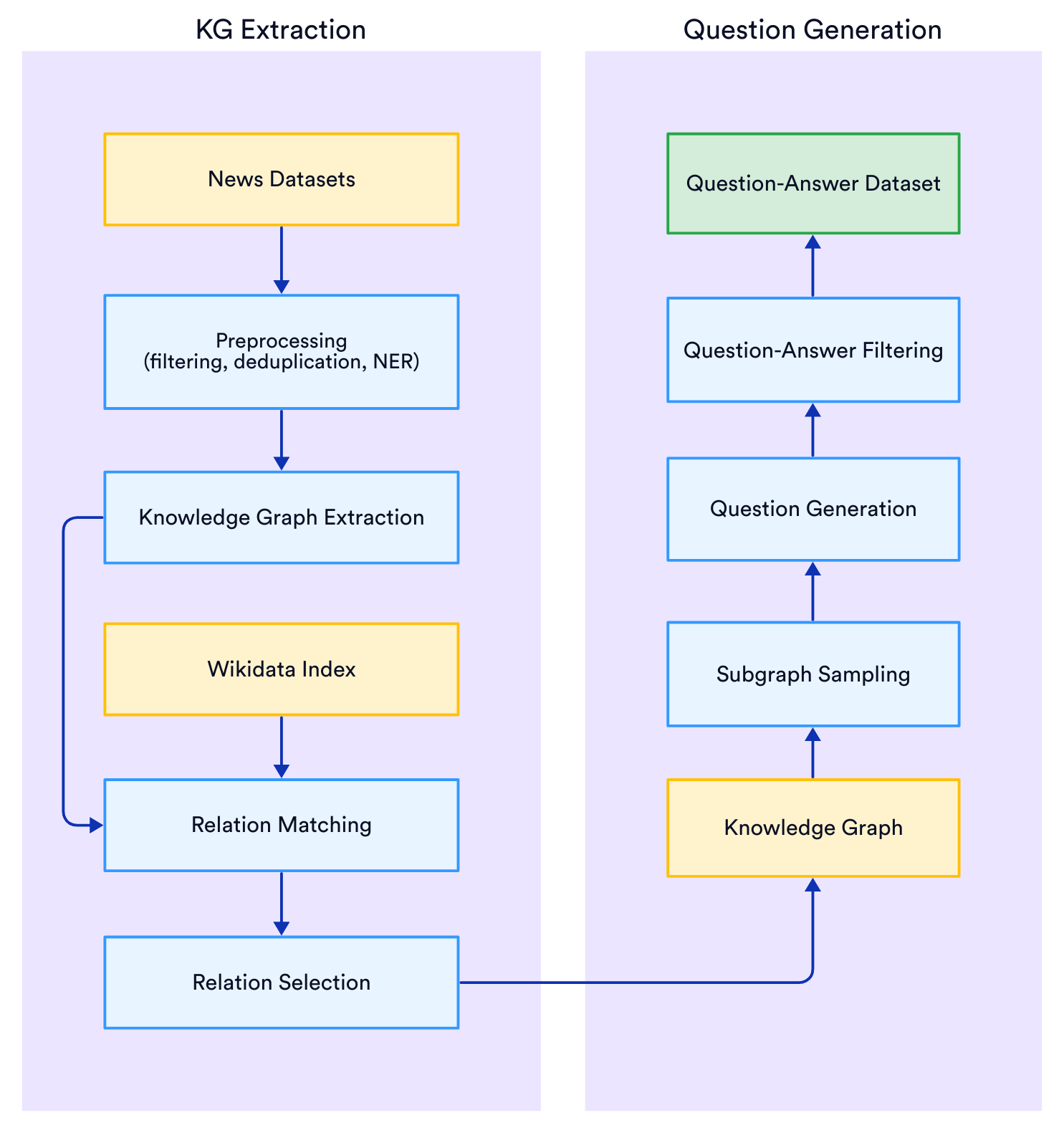}
    \caption{Architecture of the Data Generation pipeline. Before the start of the KG extraction, we perform data deduplication as the news dump could contain multiple edited versions of the same article. We preserve only the latest version of the text with the same URL. Also we extract named entities for further question filtering.}
    \label{fig:qg_pipeline}
\end{figure}

The Data Generation pipeline (see Fig.~\ref{fig:qg_pipeline}) consists of 2 main stages preceded by preliminary data preprocessing: KG Extraction and Question Generation. KG Extraction retrieves factual information from texts and preserves the most specific and fresh facts in the form of a Knowledge Graph. The Question Generation module samples subgraphs of a certain structure to generate a question-answer pair with an LLM.

\subsection{Knowledge Graph Extraction}
\label{subsec:graph_extraction}

To achieve fine-grained control over automatic question generation, we designed a Knowledge Graph (KG) Extraction module inspired by \cite{chepurova2024prompt}.
This component transforms unstructured news texts into a structured set of factual triplets that later guide question creation.

We use LLaMa 3.3 70B Instruct\footnote{\url{https://huggingface.co/meta-llama/Llama-3.3-70B-Instruct}}~\citet{grattafiori2024llama} to extract candidate factual triplets from the corpus.
Each triplet has the form (head entity – relation – tail entity), corresponding to the subject, predicate, and object in the original sentence.

The extracted entities are matched with the Russian subgraph of Wikidata~\cite{vrandevcic2014wikidata}.
For every entity name identified in the text, we query the Wikidata API to find possible matches.
The mapped entities are then vectorized using a sentence-embedding model and stored in a vector database.
To handle ambiguity, we keep the five most similar candidates for each extracted entity according to vector similarity.

To ensure consistency across triplets, the same LLaMa 3.3 70B Instruct model is used again to normalize entity and relation names.
Given the list of candidate matches from the previous step, the model selects the most appropriate canonical form while taking the full sentence context into account.
This process merges spelling variants and aliases referring to the same entity, resulting in a cleaner, unified graph structure.

Our goal is to build a graph that captures new information appearing in the latest news updates.
Therefore, we discard any triplets that exactly match existing facts in Wikidata.
Triplets absent from the knowledge base are treated as novel facts, as they are more likely to represent fresh events, and thus serve as valuable material for generating time-sensitive questions.

\subsection{Question Types}
\label{subsec:question_generation}

The question generation stage begins with subgraph extraction from the constructed Knowledge Graph. We identify all subgraphs matching one of four predefined structural templates, each representing a distinct question type~\cite{yang2024crag}:

\paragraph{Simple} These type correspond the most simple questions based a single fact mentioned in one or several texts.
They are based only on one relation from the graph: the predicate and one of the entities involved in the relation are used to compose the question, and the second entity becomes the answer. 

\begin{itemize}[noitemsep,leftmargin=1.em]
\item \textbf{relations}: \textit{(Morty Smith | voice | Keisuke Chiba)}
\item \textbf{question}: \textit{Who voiced Morty Smith?}
\item \textbf{answer}: \textit{Keisuke Chiba}
\end{itemize}

\paragraph{Set} Set questions test the RAG system's ability to align information from several texts. They are based on a one-to-many subgraphs in which the number of triplets share relation and either object or subject. The question is generated using shared entity and relation. The answer consists of all other entities in the subgraph. 

\begin{itemize}[noitemsep,leftmargin=1.em]
\item \textbf{relations}: \textit{(Ryan Otter | composed music for | Method),
(Ryan Otter | composed music for | Trigger)}
\item \textbf{question}: \textit{What projects has Ryan Otter composed music for?}
\item \textbf{answer}: \textit{Trigger, Method}
\end{itemize}

\paragraph{Multi-Hop} Multi-hop questions evaluate the system's ability to reason in a multistage manner. The corresponding subgraph is a pair of triplets, intersecting at a single entity. The question is constructed similarly to a simple question; however, the repeated entity must not be mentioned in the question. It is used as a bridge-entity, which is described in question as a reference extracted from another triplet. 

\begin{itemize}[noitemsep,leftmargin=1.em]
\item \textbf{relations}: \textit{(FAW | country of origin | China),
(FAW | number of cars sold in 2023 | 2139)}
\item \textbf{question}: \textit{In which country is the company located that sold 2139 cars in 2023?}
\item \textbf{answer}: \textit{China}
\end{itemize}

\paragraph{Conditional} Conditional questions are the extension of multi-hop questions with the same underlying subgraph of a pair of triplets, intersecting at a single entity. However, for a conditional question, both facts are used to form the question, while the repeated entity becomes an answer. 

\begin{itemize}[noitemsep,leftmargin=1.em]
\item \textbf{relations}: \textit{(Roman Miroshnichenko | performed at | M-bar),
(Roman Miroshnichenko | met with | Dmitry Dibrov)}
\item \textbf{question}: \textit{Who performed at M-bar and met with Dmitry Dibrov?}
\item \textbf{answer}: \textit{Roman Miroshnichenko}
\end{itemize}

Each selected subgraph is then passed to the language model, which generates a natural-language question–answer pair. The question is formulated as a fluent, contextually appropriate sentence, while the answer comprises one or more entities explicitly present in the subgraph.

\subsection{QA Filtering}
\label{subsec:qa_filtering}

To ensure high-quality and contextually grounded question--answer pairs, we apply a multi-stage \textbf{filtering pipeline} combining linguistic validation, entity consistency, graph correspondence check, and LLM-based judgment.

\paragraph{1. Linguistic and Structural Filtering.}
Firstly, we assess the grammatical correctness and fluency of each question using a RuRoBERTa-large model trained on the RuCoLa dataset\footnote{\url{https://huggingface.co/RussianNLP/ruRoBERTa-large-rucola}}~\cite{mikhailov2022rucola}. This step eliminates ungrammatical or poorly formed questions. 
Next, we perform Named Entity Recognition (NER) on the original source text using the \textit{Natasha} library. The generated questions and answers are checked for the presence of these entities. Samples without explicit named entities are discarded to remove trivial, knowledge-free examples. 
We further filter out overly simplistic questions by evaluating them with smaller instruction-tuned LLMs (Qwen 2.5 7B~\cite{qwen2.5} and LLaMa 3 8B~\cite{grattafiori2024llama}) without context; if a model can answer a question directly from prior knowledge, it is excluded.

\paragraph{2. Graph Correspondence Filtering.}
Each remaining QA pair is verified against the \textbf{source subgraph} used during question generation to ensure factual alignment. 
For every entity in the subgraph, we calculate the Levenshtein distance~\cite{levenshtein1966binary} between its label and the text of both the question and answer. 
Each node in the graph (entity) is assigned 2 coefficients: question presence and answer presence. It is evaluated as the scaled Levenshtein distance between the name of the entity and the closest substring from the question and answer. These values allow us to check that all entities have been mentioned correctly.

In the graphs for \textbf{Set} and \textbf{Conditional} question types, the positions of every entity are strictly determined. The algorithm averages the presence coefficients of entities implied to be in the same part of the output. If any of these values is lower than the threshold, it indicates incorrect generation. For \textbf{Simple} questions each entity can appear in both parts of the output, although the entity must be mentioned once in the question-answer pair. The presence coefficients were averaged over all entities from the subgraph, then 5\% highest and lowest values were filtered out. \textbf{Multi-Hop} questions inherit the same process for nodes having only one connection in the subgraph. The bridge entity that has two connections should not be mentioned in the model output. A high value for any of the presence coefficients for this entity demonstrates a question-type violation.

\paragraph{3. LLM-as-Judge Evaluation.}
\label{app:qa_instr}
In the final stage, we apply the \textit{LLM-as-Judge} approach using \textbf{POLLUX 7B}~\cite{martynov2025eye}, a model fine-tuned for fine-grained evaluation in Russian. 
Each QA pair is automatically rated along \textbf{eight generative criteria}:  
(1) \textit{Question literacy} (grammar and style),  
(2) \textit{Clarity},  
(3) \textit{Naturalness},  
(4) \textit{Context sufficiency} (answer can be found in the passage),  
(5) \textit{Context necessity} (question depends on the passage),  
(6) \textit{Answer correctness},  
(7) \textit{Answer uniqueness}, and  
(8) \textit{Answer literacy}.  More details on these criteria can be found in Appx.~\ref{app:qa_instr}.

Each criterion is transformed into a separate prompt with the specific scoring scale (0-2). For answer criteria (Literacy, Correctness, Uniqueness Based on Context), the prompt contains a news article, a question, and an answer; for other criteria, the answer is omitted. The example is classified as positive according to the particular criterion if the judge model assigns a rating of 1 or higher. This threshold was chosen to imitate the majority vote used in human evaluation.

To validate the reliability of using a language model as an automated evaluator for filtering generated question-answer pairs, we conducted an empirical comparison against human judgments. A random sample of 532 examples was drawn from the generated dataset and independently assessed by a panel of human annotators (with more than three annotators per example) as well as by a large language model. An example was considered positive by human annotators if
half or more of the assessors provided a positive assessment. 

\input{tables/judge_vs_human}

The comparison in Tab.~\ref{tab:llm_vs_human} reveals that the language model achieves high Precision but moderate Recall relative to the human-labeled data. This trade-off is acceptable in our setting, as the dataset contains a large volume of generated examples. In this context, precision is more critical than recall: retaining only high-quality samples is preferable, even if some potentially acceptable data are discarded. This justifies the use of the language model as an effective filter for selecting the most reliable and contextually appropriate question-answer pairs at scale.

After all filtering stages, \textbf{150 high-quality questions per category} are retained for the final benchmark dataset.

\section{Experimental Setup}
\label{sec:evaluation}

To construct our experimental RAG systems, we used the LangChain framework\footnote{\url{https://pypi.org/project/langchain/}}. All texts from the \textit{Public Texts} dataset are split into chunks of length 500 with an overlap of 100 characters. Each chunk is vectorized using the retrieval model of the evaluating RAG system with the corresponding document prefixes, and the resulting vectors are stored in a vector database. 

During the search phase, we use the prompted retrieval model to find five of the most relevant texts that match the user's query. Retrieved chunks are incorporated into a prompt provided to the LLM of the evaluated RAG system.
If the total length of the filled-in prompt exceeds the model's maximum context length, the contextual information is truncated to the required size.
To accelerate LLM inference, we utilize the vLLM framework\footnote{\url{https://github.com/vllm-project/vllm}}~\cite{kwon2023efficient}.


\subsection{Experimental Setup Details}
\label{app:exp_setup_details}

\paragraph{Embedding Model Prefixes}
\label{app:embedder_prefixes}
To vectorize questions and documents, we used embedders with the corresponding prefixes. These prefixes are shown in Tab.~\ref{tab:embedder_prefixes}.
\input{tables/emb_prefixes}

\paragraph{LLM Prompt Template}
\label{app:prompt_template}
To generate answers for the questions, we used the following template for the user message prompt:

{\small
\begin{verbatim}
``Answer the question using the provided context. 
Give me only an answer.  
<context> {context} </context>  
Question: {question}  
Answer: ''
\end{verbatim}
}

\paragraph{Model Configuration}
\label{app:model_configuration}
For serving models, we used the vLLM framework.
The model parameters used are shown in Tab.~\ref{tab:exp_setup_details}.
\input{tables/exp_setup_details}
We set \texttt{max\_new\_tokens} to 1000 for all models to limit the response length of the models.

\paragraph{Metrics}
\label{app:metrics}
The performance of retrieval is measured by 3 metrics:
\begin{itemize}
    \item \textbf{Hit Rate} measures the proportion of queries for which the relevant document appears among the top-k retrieved results.
    \item \textbf{Mean Reciprocal Rank (MRR)} evaluates ranking quality by measuring how highly the first relevant document is ranked, assigning higher scores when a relevant document appears earlier in the ranked list.
\end{itemize}

We evaluate End-to-end RAG systems with:

\begin{itemize}
\item \textbf{ROUGE-2} measures bigram overlap between the model output and the reference text, capturing local phrase-level similarity and rewarding matching adjacent word pairs.
\item \textbf{ROUGE-L}, measures the longest common subsequence between the model output and the reference text, capturing overlap in overall sentence structure.
\item \textbf{The Judge Score} is used to evaluate the overall answer quality, is calculated as the average of the automatic scores from Pollux~\footnote{\url{https://ai-forever.github.io/POLLUX/}} across multiple criteria (e.g., correctness, completeness, and relevance).
\end{itemize}

\section{Experiments}

\paragraph{Question Quality Evaluation}
\label{sec:exp_questions}

To assess the quality of the generated question-answer pairs, a human evaluation study is conducted.
Each QA pair from \textit{Sandbox Datasets} (Sec.~\ref{subsec:hf_dataset}) is independently evaluated by 3 expert annotators along the evaluation criteria from Sec.~\ref{app:qa_instr}. Annotators were asked to mark each pair as ``Good'' or ``Not Good'' with respect to each criterion. To account for potential subjectivity in judgment, we considered a QA pair to be acceptable with the majority vote. Evaluation results are provided in Tab.~\ref{tab:human_eval}.

\begin{table}[tbh!]
\centering

\begin{tabular*}{\columnwidth}{lccc}
\toprule
\textbf{Criterion} & \textbf{Apr} & \textbf{May} & \textbf{Jun} \\
\midrule
Question Literacy     & 0.96 & 0.97 & 0.99 \\
Clarity               & 0.99 & 1.00 & 1.00 \\
Naturalness           & 0.98 & 0.96 & 0.97 \\
Context Sufficiency   & 0.98 & 0.98 & 0.99 \\
Context Necessity     & 0.95 & 0.97 & 0.98 \\
\midrule
Correctness           & 0.95 & 0.92 & 0.96 \\
Uniqueness            & 0.76 & 0.78 & 0.80 \\
Answer Literacy       & 0.79 & 0.71 & 0.75 \\
\bottomrule
\end{tabular*}

\caption{
The proportion of QA pairs considered good for each dataset version and each evaluation criterion. The results establish the high quality of generated questions and significant context dependency. The answer evaluation proved the prevalence of correct answers, while the answer uniqueness is lower, so the ground truth answer can be substituted with another entity from the text. This fact exhibits the importance of LLM-as-Judge evaluation for RAG systems to avoid rephrasing influence.
}
\label{tab:human_eval}
\end{table}

\paragraph{Retrieval Evaluation}
\label{sec:exp_retrieval}
Retrieval evaluation results presented in Tab.~\ref{tab:retriever_metrics} demonstrate consistently strong performance across all evaluated retriever models. Among them, Qwen3$\textsubscript{Embedding 8B}$ and E5 Mistral$\textsubscript{7b Instruct}$ achieve the highest scores. mE5$\textsubscript{Large Instruct}$ also performs competitively. As for FRIDA, it also demonstrates strong performance but its results are slightly inferior to those of the competitors.

\begin{table}[tbh!]
\caption{Retrieval evaluation results. The best score is in bold, second best is underlined.}\label{tab:retriever_metrics}
\centering

\begin{tabular}{lcc}
\toprule
\textbf{Retriever} & \textbf{Hit Rate} & \textbf{MRR} \\
\midrule
FRIDA & 0.932 & 0.822 \\
Qwen 3$_{\textnormal{\tiny  Embedding 8b}}$ & \textbf{0.960} & \textbf{0.867} \\
E5 Mistral$_{\textnormal{\tiny   7b Instruct}}$ & \underline{0.956} & \underline{0.851} \\
mE5$_{\textnormal{\tiny   Large Instruct}}$ & 0.949 & 0.834 \\

\bottomrule

\end{tabular}


\end{table}

\paragraph{End-to-End System Evaluation}
\label{sec:exp_e2e}

\begin{table}[tbh!]
\centering
\begin{tabular*}{\columnwidth}{@{}lccc@{}}
\toprule
\textbf{LLM} & \textbf{Rouge2} & \textbf{RougeL} & \textbf{JS} \\
\midrule
\multicolumn{4}{l}{Retrieval: \textbf{FRIDA}} \\
\midrule
Gemma 3 12B it & 0.14 & 0.22 & 0.63 \\
Gemma 3 27B it & 0.14 & 0.22 & 0.64 \\
Qwen 2.5 32B & 0.09 & 0.16 & 0.62 \\
Qwen 2.5 7B & 0.08 & 0.13 & 0.57 \\
Qwen3 32B & 0.08 & 0.14 & 0.64 \\
Rudadapt Qwen 32B & 0.13 & 0.22 & 0.72 \\
\midrule
\multicolumn{4}{l}{Retrieval: \textbf{mE5$_{\textnormal{\tiny   Large Instruct}}$}} \\
\midrule
Gemma 3 12B it & 0.15 & 0.24 & 0.67 \\
Gemma 3 27B it & 0.15 & 0.24 & 0.67 \\
Qwen 2.5 32B & 0.10 & 0.18 & 0.66 \\
Qwen 2.5 7B & 0.09 & 0.15 & 0.63 \\
Qwen3 32B & 0.11 & 0.18 & 0.69 \\
Rudadapt Qwen 32B & 0.14 & 0.21 & 0.74 \\
\midrule
\multicolumn{4}{l}{Retrieval: \textbf{Qwen 3$_{\textnormal{\tiny  Embedding 8b}}$}} \\
\midrule
Gemma 3 12B it & \underline{0.16} & \textbf{0.26} & 0.71 \\
Gemma 3 27B it & \textbf{0.17} & \textbf{0.26} & 0.72 \\
Qwen 2.5 32B & 0.11 & 0.19 & 0.68 \\
Qwen 2.5 7B & 0.09 & 0.16 & 0.64 \\
Qwen3 32B & 0.11 & 0.19 & 0.71 \\
Rudadapt Qwen 32B & \underline{0.16} & \underline{0.25} & \textbf{0.82} \\
\midrule
\multicolumn{4}{l}{Retrieval: \textbf{E5 Mistral$_{\textnormal{\tiny   7b Instruct}}$}} \\
\midrule
Gemma 3 12B it & \underline{0.16} & \underline{0.25} & 0.68 \\
Gemma 3 27B it & \underline{0.16} & \underline{0.25} & 0.70 \\
Qwen 2.5 32B & 0.12 & 0.19 & 0.68 \\
Qwen 2.5 7B & 0.09 & 0.16 & 0.64 \\
Qwen3 32B & 0.11 & 0.19 & 0.71 \\
Rudadapt Qwen 32B & \underline{0.16} & \underline{0.25} & \underline{0.79} \\
\bottomrule
\end{tabular*}

\caption{End-to-end RAG-system evaluation results. Retrieval evaluation results. The judge's score (JS) is computed by averaging the results among the criteria. The best score is in bold, and the second-best score is underlined.}
\label{tab:generation_metrics}
\end{table}

The results are provided in Table~\ref{tab:generation_metrics}. Overall, the results show that classic metrics such as Rouge-L are not objective enough and do not allow evaluating all aspects of the RAG task.

First, it can be seen that the choice of the retrieval model plays a crucial role. Qwen3$_{\textnormal{\scriptsize Embedding 8B}}$ and E5 Mistral$_{\textnormal{\scriptsize7b Instruct}}$ show the strongest results.
Second, it should be noted that the general LLM ranking remains the same with every retrieval, with Rudadapt Qwen 32B heading the list by Judge Score, and Gemma 3$_{\textnormal{\scriptsize 12b it}}$ outperforming other competitors by Rouge metrics. 


In general, system scores positively characterize DRAGON as being complex enough for modern RAG-systems, allowing researchers to evaluate their capabilities at a high level. In the future, we also plan to complexify Judge Evaluation criteria, thus providing an opportunity for an adequate assessment of more advanced models than those that exist nowadays and avoiding the danger of the benchmark being solved.

\section{Conclusion}

We presented \textbf{DRAGOn}, a method to design RAG benchmark on any periodically updated document source, with it we created the dynamic benchmark for evaluating retrieval-augmented generation systems in Russian. DRAGOn is designed for real-world deployment settings by leveraging a regularly updated knowledge base and focusing on the recurrent evaluation of both retriever and generator components.
Our methodology addresses the current lack of standardized RAG evaluation tools; thus, we created a sample benchmark for the Russian language. We release the benchmark, which comprises a question generation pipeline and evaluation scripts, and launch a public leaderboard to support reproducible, transparent, and community-driven research. 
In the future, with the evolving capabilities of RAG systems, we plan to extend the benchmark by introducing  new question types, refining the LLM-as-Judge criteria. In addition, we aim to open-source previous snapshots of the evolving datasets to support reproducibility and foster further community research.

We hope DRAGOn will serve as a foundation for future work on multilingual and dynamic RAG systems.

\section*{Limitations}

While the proposed benchmark provides a valuable framework for evaluating retrieval-augmented generation (RAG) systems, several limitations should be acknowledged:

\paragraph{Source Diversity} The benchmark primarily relies on the available documents from a specific domain (news), which may not fully capture the diversity of real-world information retrieval and generation tasks. Expanding the dataset range could enhance the benchmark's applicability across different domains.

\paragraph{Language Diversity} The proposed benchmark consists entirely of Russian language documents and questions. Although the methodology itself could be easily applied to any other language, in its current state, only one language is presented.

\paragraph{Evaluation Metrics} The chosen evaluation metrics, such as ROUGE, which is essentially an n-gram precision, predominantly focus on surface-level matching. These metrics may not adequately reflect the semantic and pragmatic aspects of the generated content and have limited correlation with human judgment~\cite{deutsch2022re}. The LLM as Judge evaluation is designed to mitigate the semantic gap of n-gram-based metrics. However, the RAG benchmark requires specific criteria to capture details of system performance. Building more adapted judge models can improve the quality of the assessment.

\paragraph{Domain-Specific Challenges} RAG systems might perform differently across various domains due to domain-specific complexities and knowledge structures. The benchmark does not currently address these nuances, which could hinder its ability to generalize across distinct fields like medicine, law, or general knowledge.

\paragraph{Retriever-Generator Synergy} The interactions between retrieval and generation components are complex and dynamic. Our benchmark does not deeply explore how different configurations and synergistic interactions affect performance, possibly oversimplifying nuances that can significantly impact results.

\paragraph{Human Evaluation} The benchmark primarily relies on automated metrics, which may not align perfectly with human judgments of quality and relevance. While we acknowledge the role of human evaluation, it was not feasible to incorporate it extensively into this iteration of the benchmark.

\paragraph{Scalability and Efficiency} The computational resources required for comprehensive testing can be substantial, potentially restricting the accessibility of the benchmark to groups with extensive computational infrastructure.

\paragraph{Rapid Technological Advancements} The field of RAG systems is rapidly evolving, with new models and techniques emerging frequently. The benchmark may quickly become outdated unless regularly updated to incorporate recent advancements and methodologies.

Addressing these limitations in future work could involve developing more comprehensive, diverse datasets, incorporating a broader range of evaluation metrics, and continuously adapting the benchmark to reflect the state-of-the-art in RAG systems. Additionally, exploring detailed interactions between retrieval and generation components and integrating more human evaluation into the assessment process could provide deeper insights and improve the robustness of the benchmark.

\section*{Ethical consideration}
In developing and utilizing the retrieval-augmented generation (RAG) systems benchmark, several ethical considerations have been taken into account to ensure responsible and fair use of the technology:

\paragraph{Bias and Fairness} Given that RAG systems are influenced by the data they are trained and tested on, it's crucial to address the potential for bias in retrieval and generation processes. Our benchmark highlights these concerns by incorporating evaluation metrics that identify and measure biases in model outputs. Future iterations aim to include datasets specifically designed to stress-test and mitigate bias.

\paragraph{Data Privacy} The use of real-world datasets in RAG systems poses privacy risks, particularly concerning personally identifiable information (PII). We ensure that datasets included in the benchmark are sourced following strict privacy regulations and guidelines, and we encourage the anonymization of any PII to safeguard user privacy.

\paragraph{Content Quality and Misinformation} RAG systems can potentially generate or propagate misinformation if not properly managed. Our benchmark assesses models on their ability to produce accurate and reliable content, and we emphasize the importance of retrieval sources that are reputable and verifiable to minimize risks associated with misinformation.

\paragraph{Transparency and Explainability} Understanding the decision-making process of RAG systems is critical for trust and accountability. The benchmark encourages the development of models that offer insights into their retrieval and generation processes, promoting transparency and explainability.

\paragraph{Unintended Consequences} The application of RAG systems can have unintended societal impacts, such as fostering dependency on AI for decision-making or influencing cultural narratives. Researchers and developers are encouraged to consider these broader implications and involve interdisciplinary perspectives in assessing the impact of their systems.

\paragraph{Access and Inequality} High computational demands of RAG systems can exacerbate the divide between well-resourced organizations and smaller entities or individuals. Our benchmark advocates for the creation of more efficient models that democratize access and enable wider participation in developing and utilizing RAG technology.

\paragraph{Responsible Usage} Educating users and stakeholders about the capabilities and limitations of RAG systems is vital to prevent misuse. Our research promotes guidelines and best practices to ensure that these technologies are used responsibly and ethically.

By acknowledging and addressing these ethical considerations, our aim is to contribute positively to the development and deployment of retrieval-augmented generation systems, ensuring they serve society in a beneficial and responsible manner. Future work will continue to refine these frameworks to address emerging ethical challenges as the field evolves.

\paragraph{Error Analysis} A further limitation of our current benchmark release is the lack of a systematic error analysis of model failures. While we report aggregate retrieval and generation scores, we do not yet provide a fine-grained breakdown of common failure modes (e.g., retrieval misses vs. ranking issues, incomplete evidence aggregation in multi-hop questions, hallucinations under partially relevant context). Such analysis is important both for interpreting leaderboard progress and for understanding whether improvements come from better retrieval, better grounding/faithfulness, or exploiting dataset artifacts. In future work, we plan to add structured error taxonomies and identify systematic weaknesses of evaluated RAG pipelines.

\paragraph{AI-assistants Help} We improve and proofread the text of this article using Writefull assistant integrated in Overleaf (Writefull's/Open AI GPT models) and GPT-4o\footnote{\url{https://chatgpt.com}}, Grammarly\footnote{\href{https://app.grammarly.com/}{https://app.grammarly.com/}} to correct grammatical, spelling, and style errors and paraphrase sentences. We underline that these tools are used strictly to enhance the quality of English writing, in full compliance with the ACL policies on responsible use of AI writing assistance.
Nevertheless, some segments of our publication can be potentially detected as AI-generated, AI-edited, or human-AI-generated.

\section*{Acknowledgments}

We would like to express our deep appreciation to Ivan Bondarenko for his contribution to our pipeline and generous support of this work. This research partially done by A.F. is an output of a research project implemented as part of the Basic Research Program at the National Research University Higher School of Economics (HSE University).

\bibliography{ecir_custom}

\appendix
\label{app:appendix}
\section{News Data Sources}
\label{app:sources}
For dataset formation, we rely on content from several well-established Russian news websites\footnote{All data is used in full compliance with legal requirements and ethical standards, under a formal agreement with Rambler. The collection process ensures respectful use of content without infringing on the rights of publishers or individuals.}:
\begin{itemize}[nosep]
  \item \url{blog.okko.tv},
  \item \url{daily.afisha.ru},
  \item \url{lenta.ru},
  \item \url{letidor.ru},
  \item \url{moslenta.ru},
  \item \url{motor.ru},
  \item \url{quto.ru},
  \item \url{tass.ru},
  \item \url{gazeta.ru},
  \item \url{ria.ru},
  \item \url{rg.ru}.
\end{itemize}

\section{Leaderboard Overview}
\label{sec:app_lb}
\begin{figure*}[tbh!]
    \centering
    \includegraphics[width=0.8\linewidth]{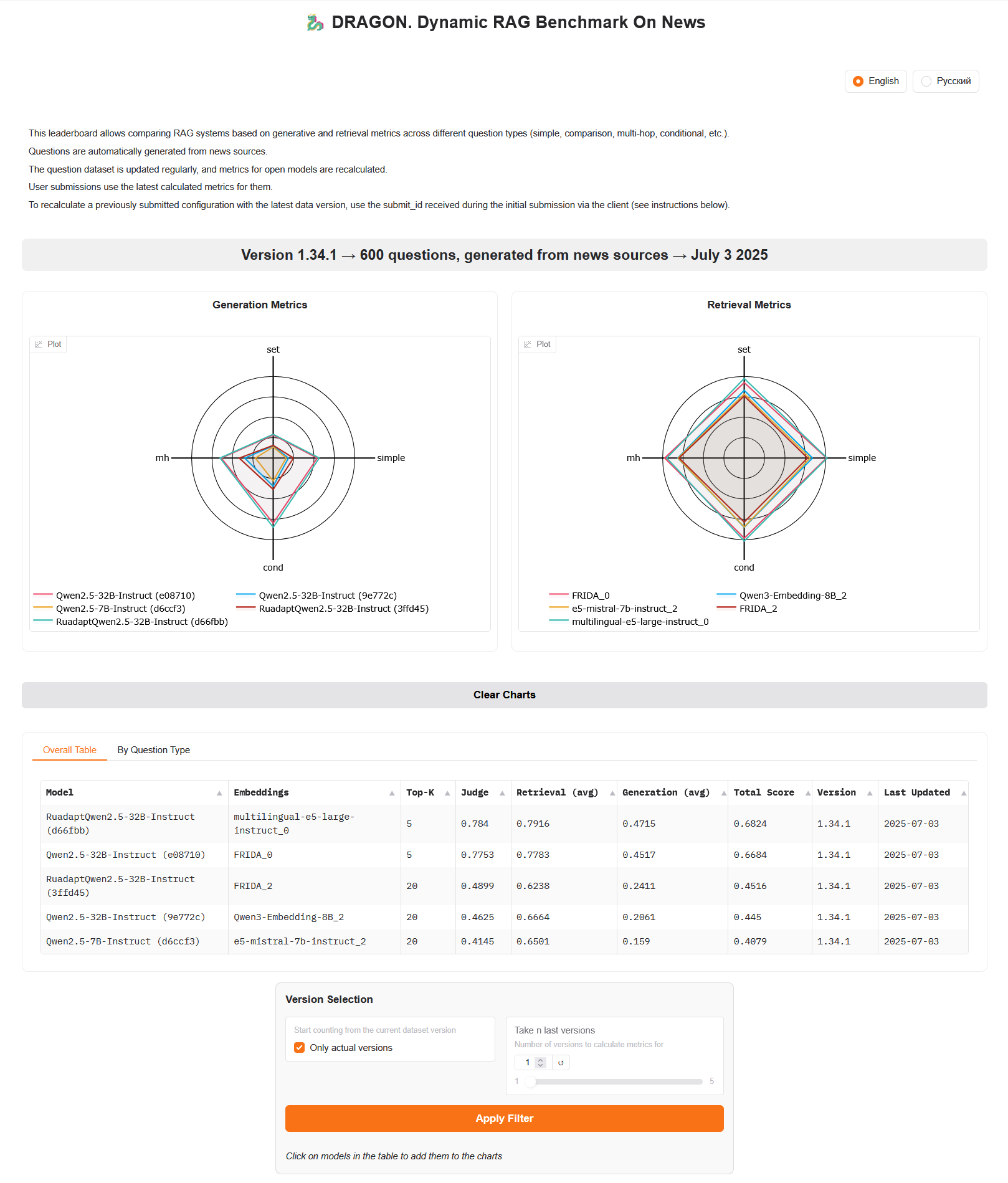}
    \caption{Leaderboard interface.}
    \label{fig:leaderboard}
\end{figure*}

Fig.~\ref{fig:leaderboard} shows an overview of the leaderboard.

\section{Baseline Details}
We evaluate open-source LLMs within 70B size\footnote{The size limit is introduced to ensure the feasibility of multi-model evaluation under the compute budgets.} which score best on the MERA benchmark\footnote{\url{https://mera.a-ai.ru/en/text/leaderboard}, valid for July 1, 2025.}~\cite{fenogenova2024mera} (see Tab.~\ref{tab:llm_desc} for their description) and several popular embedding models which show strong results on the retrieval task on ruMTEB or Multilingual MTEB\footnote{\url{https://huggingface.co/spaces/mteb/leaderboard} valid for July 1, 2025.}~\cite{snegirev2024russian, enevoldsen2025mmteb}. 

\label{app:baseline_detail}

\input{tables/LLM_description}

\section{Question-Answer Evaluation Criteria}
\label{app:qa_instr}

This section describes the question-answer evaluation criteria used on the final question filtering stage. These criteria were developed to assess the general quality and naturalness of the question, its context dependence, and the correctness of the answer.
The same set of criteria is used for manual annotation.

\paragraph{Question Literacy} \textit{Does the question exhibit correct grammar, spelling, and punctuation?}
This criterion assesses the linguistic quality of the question. A well-formed question should be free of typographical errors, contain appropriate punctuation, and follow standard grammatical rules. Additionally, the phrasing should align grammatically with the surrounding context, ensuring the question does not feel syntactically out of place.

\paragraph{Question Clarity} \textit{Is the intent of the question clear and unambiguous?}
This criterion evaluates how easily a reader can understand what information is being requested. The question should be interpretable either based on the provided context or general knowledge, without requiring additional clarification. Vague, overly broad, or logically inconsistent questions should be penalized.

\paragraph{Question Naturalness} \textit{Does the question sound like it could have been written by a human?}
This assesses whether the question appears natural and contextually appropriate. It should avoid signs of being artificially generated such as unnatural phrasing, rigid templates, or repetitive structures. A natural question should feel relevant and plausible within the discourse of the text.

\paragraph{Context Sufficiency} \textit{Can the answer to this question be found entirely within the provided context?}
This criterion determines whether the context passage contains enough information to answer the question. A question should not require external knowledge or assumptions unless that knowledge is very general or trivial. Questions with answers that are clearly present and verifiable in the text should receive high marks.

\paragraph{Context Necessity} \textit{Is the provided context necessary to answer the question?}
This evaluates whether the question meaningfully engages with the context. Ideal questions should be context-dependent, meaning they cannot be accurately answered without access to the specific passage. Generic or overly broad questions that could be answered independently of the text (without specialized knowledge) are discouraged.

\paragraph{Answer Literacy}
\textit{Is the answer written in a grammatically correct and readable manner?}
This criterion checks for the overall linguistic quality of the answer. It should be free from spelling mistakes, awkward constructions, or inconsistent grammatical structure.

\paragraph{Answer Correctness}
\textit{Is the answer factually correct and appropriate for the given question?}
This criterion gauges the accuracy of the generated answer. It should contain all possible entities that can be mentioned in the answer without omitting any necessary details.

\paragraph{Answer Uniqueness Based on Context}
\textit{Is this the only plausible answer that can be given based on the text?}
This checks whether the answer is uniquely determined by the information in the context. If the passage contains multiple plausible answers or if ambiguity remains, this checkbox should not be selected. Ideal answers should be both correct and exclusive given the text.

\section{Human Evaluation Interface}
\label{app:eval_interface}
A screenshot of a system used for human evaluation is presented in Fig.~\ref{fig:judge_criteria}.

\begin{figure*}[tbh!]
    \centering
    \begin{subfigure}[b]{0.8\textwidth}
        \centering
        \includegraphics[width=\linewidth]{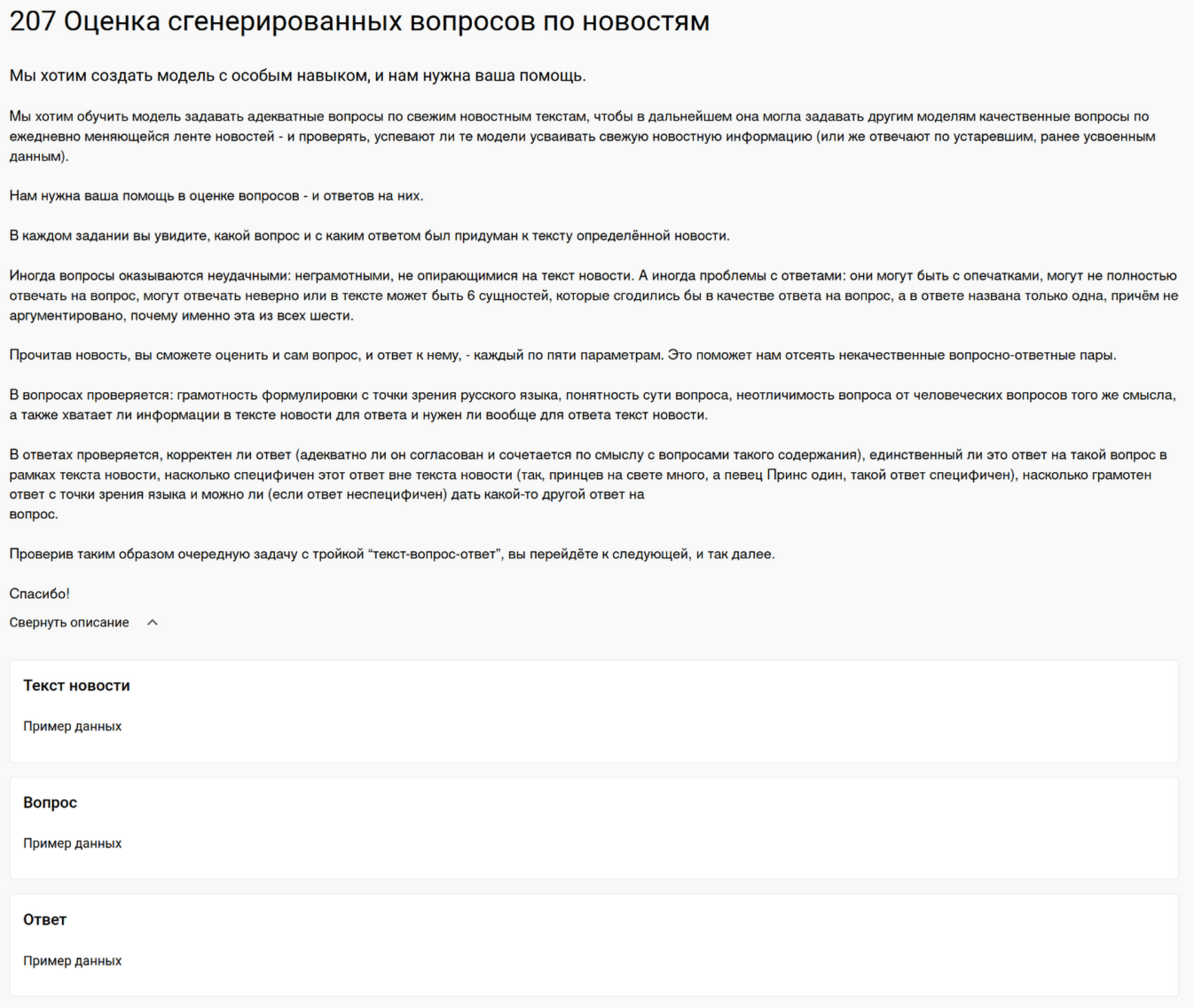}
        \caption{Interface part 1}
    \end{subfigure}
    \vfill
    \begin{subfigure}[b]{0.8\textwidth}
        \centering
        \includegraphics[width=\linewidth]{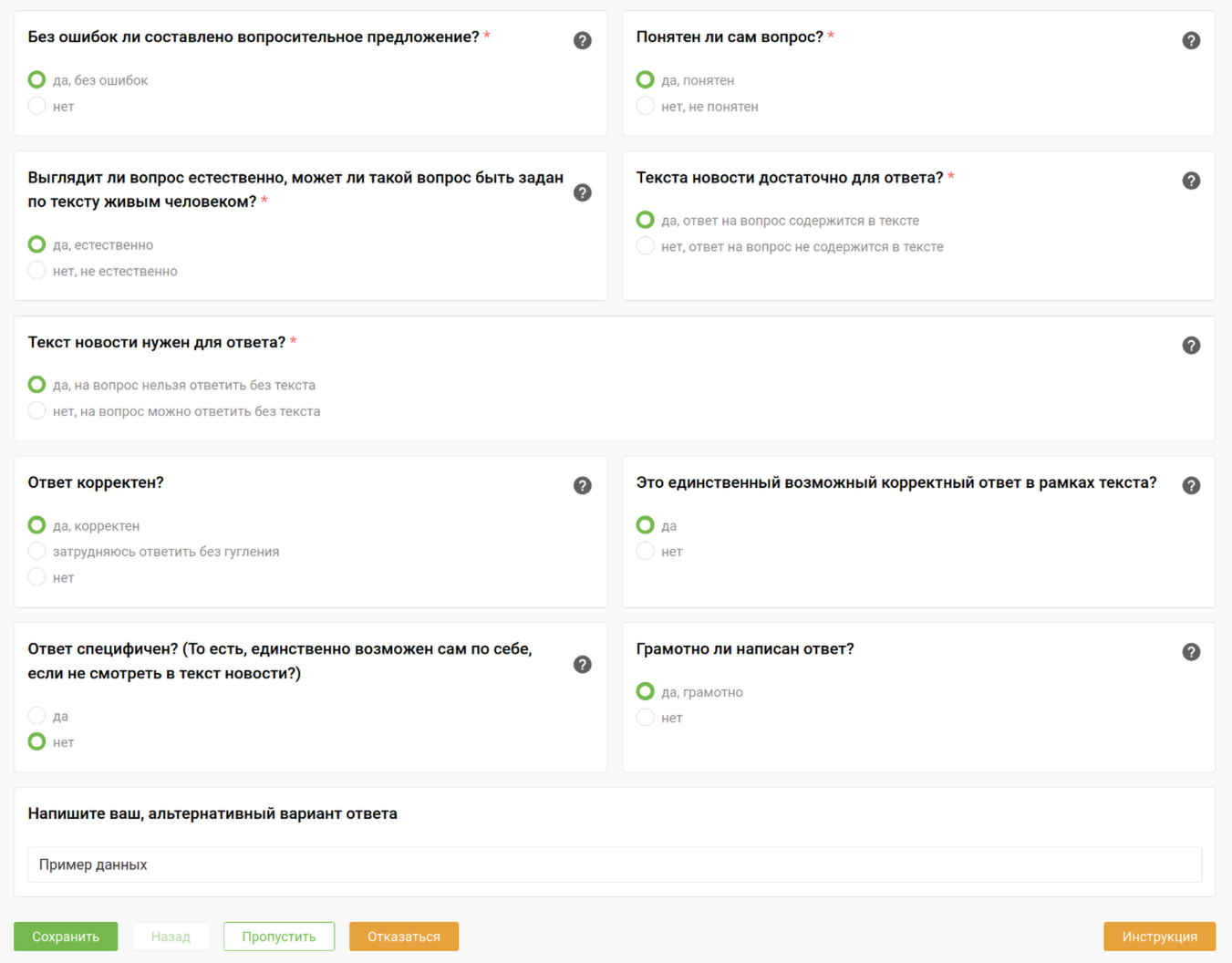}
        \caption{Interface part 2}
    \end{subfigure}
    \caption{Human evaluation system interface.}
    \label{fig:judge_criteria}
\end{figure*}

\section{LLM-as-Judge RAG Evaluation Criteria}
\label{app:judge_detailed}

This section provides a detailed description of the LLM-as-Judge criteria used to evaluate RAG systems. To build a comprehensive and interpretable set of metrics, Evaluation Targets provided by~\citet{yu2024evaluation} are utilized. For each Evaluation Target, we select several criteria from the POLLUX set of criteria:

\begin{itemize}
    \item \textbf{Answer Relevance.} Measures the alignment between the generated response and the content of the initial query.
    \begin{itemize}
        \item \textit{Absence of unnecessary details. (Fluff)} The LLM's output is relevant and do not contain fluff.
    \end{itemize}
    \item \textbf{Faithfulness.} Estimates the quality of the information extraction from retrieved documents.
    \begin{itemize}
        \item \textit{Consistency with real-world facts.} The LLM's output does not contain factual errors.
        \item \textit{Correctness of results.} The LLM extracted correct information from the text.
    \end{itemize}
    \item \textbf{Correctness} Measures the accuracy of the generated response by comparing it to the ground truth response.
    \begin{itemize}
        \item \textit{Completeness.} The answer is complete and reaches the goal.
        \item \textit{Factual accuracy.} The LLM correctly reproduced the necessary facts and their related context.
        \item \textit{Preserving the main idea and details of the original.} The LLM preserves details and main idea.
    \end{itemize}
\end{itemize}

The fine-grained set of metrics allows for comparing the RAG systems more precisely and improves interpretability. Fig.~\ref{fig:judge_criteria} provides a comparison of different RAG systems built on the basis of different variants of the Qwen 2.5 model combined with FRIDA and Qwen 3 Embedding 8B retrieval models. The results clearly demonstrate that larger language models yield higher-quality responses across all criteria, while the Absence of unnecessary details criterion results are similar for all combinations. Additionally, systems using Qwen3 8B embeddings consistently outperform those using FRIDA, highlighting the critical role of retrieval quality in end-to-end RAG performance. These findings emphasize that both the generative and retrieval components contribute significantly to final system effectiveness.

\begin{figure*}[tbh!]
    \centering
    \includegraphics[width=0.8\linewidth]{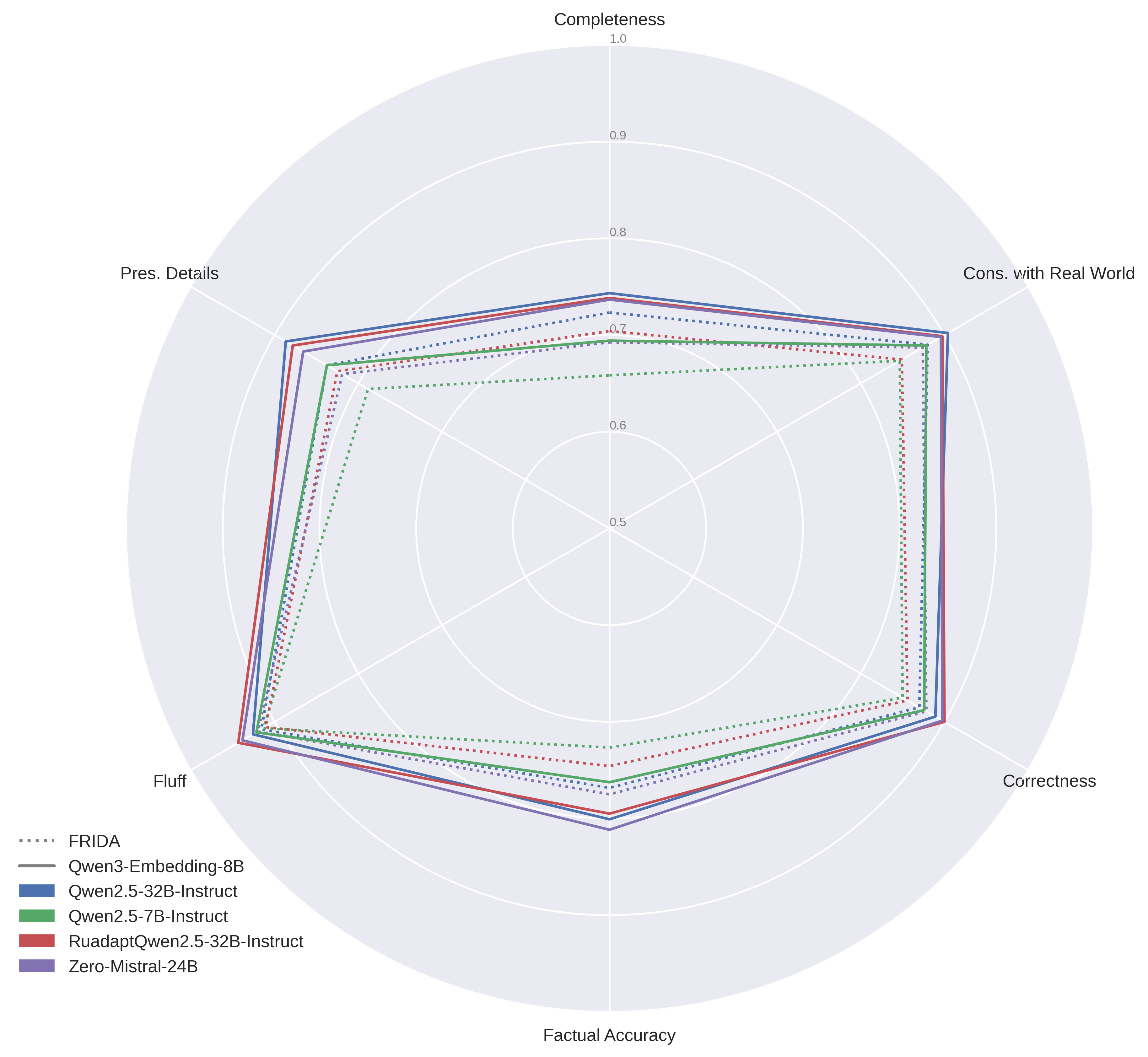}
    \caption{Detailed analysis of the RAG system performance from Tab.~\ref{tab:generation_metrics} along the separate criteria.
    }
    \label{fig:judge_criteria}
\end{figure*}

\end{document}

%% file: tables/judge_vs_human.tex
\begin{table}[tb!]
    \centering
    
    \begin{tabular*}{\columnwidth}{@{}lrr@{}}
        \toprule
        \textbf{Criterion} & \textbf{Precision} & \textbf{Recall} \\
        \midrule
        Question Literacy & 0.96 & 0.99 \\
        Question Clarity & 0.99 & 0.62 \\
        Question Naturalness & 0.96 & 0.52 \\
        Context Sufficiency & 0.94 & 0.71 \\
        Context Necessity & 0.93 & 0.95 \\
        Answer Correctness & 0.95 & 0.82 \\
        Answer Uniqueness & \multirow{2}{*}{0.85} & \multirow{2}{*}{0.78}\\
        based on context & & \\
        Answer Literacy & 0.91 & 0.97 \\
        \bottomrule
    \end{tabular*}
    \caption{Comparison of the automatic metrics and manual evaluation results. The model achieves high \textbf{precision} but moderate \textbf{recall} relative to human evaluation. 
This trade-off is desirable in our setting, where maintaining dataset reliability is more important than exhaustive coverage. 
Retaining only high-confidence samples ensures that the resulting benchmark consists of the most coherent, contextually valid, and factually grounded question-answer pairs.}
\label{tab:llm_vs_human}
\end{table}

%% file: tables/emb_prefixes.tex
\begin{table*}[h!]
	\centering
	\footnotesize
	\begin{tabularx}{
		\textwidth}{l>{
		\raggedright\arraybackslash}X>{\raggedright\arraybackslash}X}
		\toprule
		\textbf{Model}                                        & \textbf{Query prefix}                                                                        & \textbf{Text prefix} \\
		\midrule
		FRIDA                                                 & search\_query:                                                                               & search\_document:    \\
		E5 Mistral$_{\textnormal{\scriptsize   7b Instruct}}$ & Instruct: Given a web search query, retrieve relevant passages that answer the query. Query: & X                    \\
		Qwen 3$_{\textnormal{\scriptsize  Embedding 8b}}$     & Instruct: Given a web search query, retrieve relevant passages that answer the query. Query: & X                    \\
		mE5$_{\textnormal{\tiny   Large Instruct}}$           & Instruct: Given a web search query, retrieve relevant passages that answer the query. Query: & X                    \\
		\bottomrule
	\end{tabularx}
	\caption{Embedder configurations: query and text prefixes}
	\label{tab:embedder_prefixes}
\end{table*}

%% file: tables/exp_setup_details.tex
\begin{table}
    \centering
   
        \begin{tabular}{lcc}
            \toprule
            \textbf{Model} & \textbf{TP} & \textbf{ML} \\
            \midrule
            Qwen 2.5$_{\textnormal{\scriptsize 32b Instruct}}$    & 4 & 32768 \\
            Qwen 2.5$_{\textnormal{\scriptsize 7b Instruct}}$      & 1 & 32768 \\
            Ruadapt Qwen$_{\textnormal{\scriptsize 32b Instruct}}$ & 4 & 32768 \\
            Qwen 3$_{\textnormal{\scriptsize 32B}}$ & 4 & 32768 \\
            Gemma 3$_{\textnormal{\scriptsize 12b it}}$ & 1 & 131072 \\
            Gemma 3$_{\textnormal{\scriptsize 27b it}}$ & 4 & 131072 \\
            \bottomrule
        \end{tabular}
    \caption{Model configurations. \textbf{TP} stands for tensor parallel size, \textbf{ML} for maximal context length.}
    \label{tab:exp_setup_details}
\end{table}

%% file: tables/LLM_description.tex

\begin{table*}[]
\centering
\renewcommand{\arraystretch}{1.05}
\resizebox{\linewidth}{!}{
\begin{tabular}{@{}llll@{}}
\toprule
\textbf{Model}            & \textbf{Size} & \textbf{Hugging Face Hub link}                                                            & \textbf{Citation}                                                                                \\ \midrule
Qwen 2.5$_{\textnormal{\scriptsize 7b Instruct}}$      & 32B                 & \href{https://huggingface.co/Qwen/Qwen2.5-32B-Instruct}{Qwen/Qwen2.5-32B-Instruct}             & \cite{qwen2.5}                                   \\
Qwen 2.5$_{\textnormal{\scriptsize 32b Instruct}}$    & 32B                 & \href{https://huggingface.co/Qwen/Qwen2.5-32B-Instruct}{Qwen/Qwen2.5-32B-Instruct}             & \cite{qwen2.5}                                                                    \\
Ruadapt Qwen$_{\textnormal{\scriptsize 32b Instruct}}$ & 32B                 & \href{https://huggingface.co/msu-rcc-lair/RuadaptQwen-32B-instruct}{msu-rcc-lair/RuadaptQwen-32B-instruct} & \cite{tikhomirov2024facilitating}\\

Qwen 3$_{\textnormal{\scriptsize 32B}}$ & 32B & \href{https://huggingface.co/Qwen/Qwen3-32B}{Qwen/Qwen3-32B}& \cite{qwen3technicalreport} \\

Gemma 3$_{\textnormal{\scriptsize 12b it}}$               & 12B                & \href{https://huggingface.co/google/gemma-3-12b-it}{google/gemma-3-12b-it}                 & \cite{gemma_2025}                                                              \\
Gemma 3$_{\textnormal{\scriptsize 27b it}}$               & 27B                 & \href{https://huggingface.co/google/gemma-3-27b-it}{google/gemma-3-27b-it}                 & \cite{gemma_2025}   \\

\bottomrule                                                        
\end{tabular}
}
\caption{The evaluated model description. Instruct models are marked with the corresponding suffix.}
\label{tab:llm_desc}
\end{table*}

\begin{table*}[]
\centering
\renewcommand{\arraystretch}{1.05}

\begin{tabular}{@{}llll@{}}
\toprule
\textbf{Model}            & \textbf{Size} & \textbf{Hugging Face Hub link}                                                            & \textbf{Citation}                                                                                \\ \midrule

FRIDA       &  823M                  & \href{https://huggingface.co/ai-forever/FRIDA}{ai-forever/FRIDA}  & --                                                          \\
Qwen 3$_{\textnormal{\scriptsize  Embedding 8b}}$        & 8B                  & \href{https://huggingface.co/Qwen/Qwen3-Embedding-8B}{Qwen/Qwen3-Embedding-8B}  & \cite{qwen3embedding}                                                          \\
E5 Mistral$_{\textnormal{\scriptsize   7b Instruct}}$        & 7B                  & \href{https://huggingface.co/intfloat/e5-mistral-7b-instruct}{intfloat/e5-mistral-7b-instruct}  & \cite{wang2023improving}                                                          \\

mE5$_{\textnormal{\scriptsize   Large Instruct}}$ & 560M & \href{https://huggingface.co/intfloat/multilingual-e5-large-instruct}{intfloat/multilingual-e5-large-instruct} & \cite{wang2024multilingual} \\
\bottomrule                                                        
\end{tabular}

\caption{The evaluated retriever description. Instruct models are marked with the corresponding suffix.}
\label{tab:retriever_desc}
\end{table*}